\documentclass{article}
\usepackage{iclr2027_conference,times}
\iclrfinalcopy % Show author names and suppress review line numbers for this preprint.
\usepackage{amsmath,amssymb,booktabs,graphicx,array,multirow,tikz,xcolor}
\usepackage{needspace}
\usepackage{float}
\usepackage{hyperref}
\usepackage{url}
\hypersetup{hidelinks,pdfauthor={Mingxiang Shen, Jinhao Li, Jiaming Xu, Yibo Han, Guohao Dai, Yijia Zhang, Ningyi Xu},pdfsubject={Research preprint},pdftitle={TRUER: Test-Time Reward Correction Using an Error-Aware Rectifier for Reinforcement Learning}}
\makeatletter
\let\inputtable\@@input
\makeatother
\newcommand{\ind}{\mathbb{1}}

\title{TRUER: Test-Time Reward Correction\\ Using an Error-Aware Rectifier\\ for Reinforcement Learning}
\author{%
\textbf{Mingxiang Shen\quad Jinhao Li\quad Jiaming Xu\quad Yibo Han}\\
\textbf{Guohao Dai\quad Yijia Zhang\quad Ningyi Xu}\\[0.5ex]
\normalfont Shanghai Jiao Tong University
}
\begin{document}
\maketitle
\lhead{Preprint}

\begin{abstract}
Test-time reinforcement learning enables language models to adapt using rewards derived from their own answers. However, majority voting can reinforce incorrect solutions even when correct alternatives are available. Our training audit shows that false positives---incorrect responses receiving positive rewards---dominate reward errors. We focus on a practical regime where ground-truth rewards are difficult to obtain at scale, while a limited labeled source set is available. We introduce TRUER, a framework that uses limited supervision to train a reward rectifier that distinguishes correctness from consensus. To achieve this goal, we train the rectifier with an emphasis on incorrect majorities and correct minorities, targeting cases where consensus and correctness disagree. After that, we freeze the rectifier and use it to correct rewards during subsequent test-time adaptation. By improving reward accuracy and thereby model performance, TRUER consistently outperforms naive TTRL in both metrics across all evaluated configurations.
\end{abstract}

\section{Introduction}
Reliable rewards are central to reinforcement learning for language model reasoning \citep{shao2024deepseekmath,deepseekai2025r1}. Yet ground-truth rewards can be difficult to obtain at scale. Test-time reinforcement learning (TTRL) addresses this challenge by deriving pseudo-rewards from the model's own responses \citep{zuo2025ttrl}. A common approach samples several solutions and uses the most frequent answer as a training target, extending the idea of self-consistency from inference to learning \citep{wang2023selfconsistency}.

But consensus does not ensure correctness. Even when a correct answer appears among the model's 32 sampled responses, majority voting can favor an incorrect answer, as illustrated by the gap between Pass@32 and Majority@32 in Figure~\ref{fig:motivation}(a). Majority-based rewards can then reinforce an incorrect majority while withholding reward from a correct minority. The challenge is therefore to distinguish correct answers from false consensus and use them to guide subsequent learning.

Ground-truth answers provide direct supervision for this distinction, but obtaining them at scale for target problems can be costly. Prior work uses limited supervision to assess pseudo-label quality \citep{li2024semireward} and guide reinforcement learning on unlabeled problems \citep{yang2025trapo}. We therefore focus on a practical regime with limited labeled source data, disjoint from the target adaptation questions, while target ground-truth rewards remain unavailable. We investigate how this source supervision can help correct erroneous consensus without accessing target ground-truth answers.

To this end, we introduce \textbf{T}est-Time \textbf{R}eward Correction \textbf{U}sing an \textbf{E}rror-Aware \textbf{R}ectifier (\textbf{TRUER}), which turns limited source supervision into reusable reward correction. The rectifier is trained on candidate solutions organized by both correctness and consensus membership, with emphasis on incorrect majorities and correct minorities. After training, it is frozen. During test-time adaptation, the rectifier verifies the leading answer, checks a second-ranked alternative after explicit rejection, and abstains when neither provides a reliable target. This design aims to improve reward accuracy while retaining useful learning signals.

\begin{figure}[t]
\centering
\includegraphics[width=\linewidth]{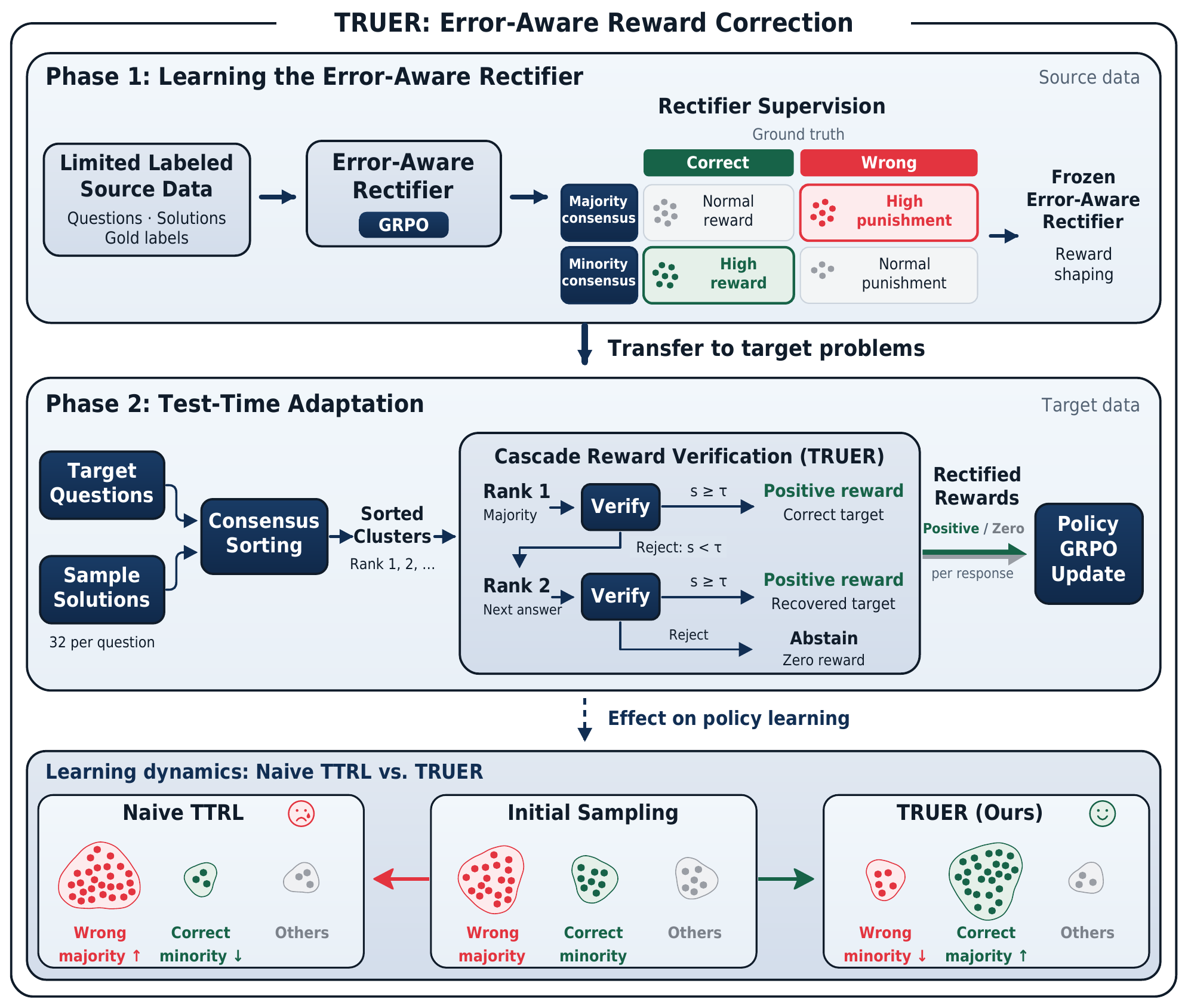}
\caption{\textbf{TRUER overview and learning dynamics.} Top: limited source labels train an error-aware rectifier, which is then frozen to correct rewards during test-time adaptation. Bottom: starting from the same sampled answer distribution, naive TTRL can reinforce a wrong majority, whereas TRUER can recover a correct minority and support policy learning.}
\label{fig:overview}
\end{figure}

Our experiments evaluate reward quality and final model performance across several mathematical reasoning configurations, comparing TRUER with naive TTRL and an oracle reference trained with ground-truth rewards using GRPO. Our contributions are threefold.
\begin{itemize}
\item \textbf{Reward reliability in majority-vote TTRL.} We analyze the reliability of majority-based rewards in TTRL. Despite high Pass@32, consensus can miss correct candidates; our training audit further shows that reward accuracy does not improve consistently and that false positives dominate reward errors.
\item \textbf{Limited-supervision reward rectification.} We introduce \textbf{TRUER}, which emphasizes incorrect majorities and correct minorities when training a rectifier on a disjoint labeled source set. The frozen rectifier rejects incorrect consensus, recovers alternatives, and abstains when needed, correcting rewards without querying target ground truth.
\item \textbf{Empirical evaluation and analysis.} Across all evaluated configurations, TRUER improves reward accuracy and final model performance over naive TTRL, narrowing the gap to GRPO trained with ground-truth rewards. Reward audits, training trajectories, and ablations further examine how reward correction changes supervision and supports policy learning.
\end{itemize}

\section{Related Work}
\label{sec:related}
\paragraph{Test-time learning with self-generated rewards.}
TTRL \citep{zuo2025ttrl} uses majority voting to estimate labels for reinforcement learning. Its ``lucky hit'' effect can maintain high reward accuracy because incorrect answers differing from an incorrect pseudo-label still receive correct zero rewards. However, consensus does not ensure correctness, and rewarding incorrect consensus may reinforce the model's existing reasoning errors.

\paragraph{Pseudo-label selection through consistency and reflection.}
Self-Harmony \citep{wang2025selfharmony} uses a reframer to reformulate questions, then selects answers receiving consistent support across original and reframed versions. SR-TTRL \citep{wu2026srttrl} instead groups and summarizes candidate solutions, then prompts the model to compare their reasoning and select a pseudo-label. Meanwhile, SCRL \citep{yan2026scrl} restricts positive rewards to sufficiently strong consensus and uses entropy-gated negative feedback to suppress unlikely answers. These methods derive corrective signals from the model's answer distributions or self-assessment. When these signals share the model's reasoning errors, they may provide insufficient evidence to reject incorrect majorities or recover correct minorities.

\paragraph{Verification training with consensus-derived supervision.}
CoVerRL \citep{pan2026coverrl} jointly trains generation and verification: consensus-derived contrastive examples supervise the verifier, whose judgments then filter pseudo-labels. Similarly, TTRL-CoCoV \citep{li2026cocov} uses high-confidence majority answers to train verification and applies the resulting judgments to select pseudo-labels for low-confidence problems. Although this interaction can improve supervision, verification targets still depend on model consensus. Consequently, incorrect consensus may propagate into verification training, reinforcing existing biases and limiting corrective feedback.

\paragraph{Reward correction with limited supervision.}
CARE \citep{wang2026care} selects problems for annotation using estimated reliability and correction value, but requires ground-truth access during adaptation. Despite labeled supervision, its acquisition classifiers use policy-derived features that may reflect model biases. Our method instead trains a reward rectifier on a labeled source set disjoint from the target problems, emphasizing incorrect majorities and correct minorities. The rectifier is then frozen and reused to correct rewards without acquiring ground-truth labels during target adaptation.

\section{When Consensus Fails to Provide Reliable Supervision}
\label{sec:motivation}
\subsection{Correct Solutions Can Be Missed by Consensus}
\begin{figure}[t]
\centering
\includegraphics[width=\linewidth]{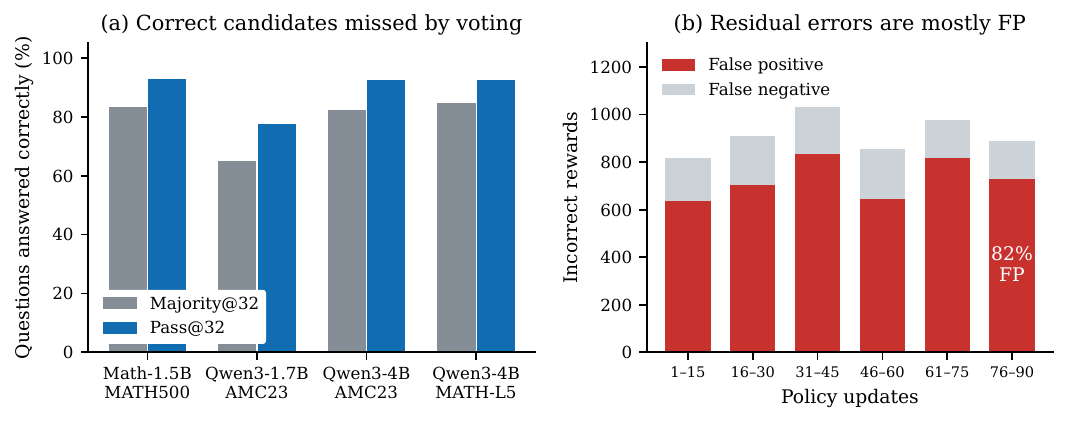}
\caption{\textbf{Available correct answers do not guarantee reliable rewards.} (a) Pass@32 and Majority@32 use the same candidate pool at each naive-TTRL endpoint. The gap represents correct candidates missed by voting. (b) In the Qwen2.5-Math-1.5B/MATH500 training audit, false positives dominate reward errors across disjoint training windows, including late training.}
\label{fig:motivation}
\end{figure}

Candidate coverage and answer selection are distinct aspects of inference-time scaling \citep{brown2024monkeys}. For each question, Pass@32 measures whether a correct answer appears in a fixed pool of sampled responses, whereas Majority@32 measures whether voting selects it. Their difference identifies questions on which a correct candidate is available but overlooked. Figure~\ref{fig:motivation}(a) shows this gap across the inspected naive-TTRL endpoints. It is particularly pronounced for the smaller Qwen3 model on AMC23. Correct solutions are reachable, but answer frequency does not reliably identify them.

This gap motivates checking correctness before turning a popular answer into supervision, consistent with the logical-minority diagnosis of \citet{wu2026srttrl}.

\subsection{False Positives Dominate Residual Reward Errors}
Let $a_i$ be the answer extracted from response $y_i$, $m$ the most frequent answer, and $a^*$ the ground-truth answer. We use majority to mean the largest answer cluster, which need not contain more than half the responses. Majority and correctness rewards are
\begin{equation}
 r_i^{\mathrm{mv}}=\ind[a_i\equiv m],\qquad
 r_i^*=\ind[a_i\equiv a^*],
 \label{eq:rewards}
\end{equation}
where $\equiv$ denotes mathematical equivalence. A false positive (FP) rewards an incorrect response; a false negative (FN) withholds reward from a correct response. Overall reward accuracy measures agreement between these binary rewards and true correctness.

Our MATH500 audit shows that continued training does not consistently improve reward accuracy. Across consecutive 15-update windows, accuracy fluctuates and ends at 94.21\%, below the initial 94.67\%. Meanwhile, false positives dominate reward errors in every window, accounting for 82\% in the final window (Figure~\ref{fig:motivation}(b)). Incorrect answers outside the selected cluster still receive the correct zero reward, keeping aggregate accuracy high despite these erroneous positive rewards.

\subsection{Implications for Reward Correction}
With group-relative policy optimization (GRPO), a rewarded response receives positive outcome advantage when its group contains mixed rewards \citep{shao2024deepseekmath}. A wrong majority can therefore be reinforced while a correct minority is discouraged. Constant-reward groups have no such outcome contrast.

For example, suppose that 18 of 32 responses share an incorrect answer, 10 give the correct answer, and the remaining four give other incorrect answers. Majority voting assigns reward 1 to the 18 incorrect responses and reward 0 to the correct minority. Under GRPO, this reward contrast can increase the probability of the incorrect answer, making it more likely to dominate subsequent samples. Repeated updates can thus reinforce the wrong majority and make the correct answer harder to recover.

A useful correction policy must address both sides of this failure. Rejecting an incorrect majority removes false-positive rewards; accepting a correct alternative restores a learning target. Rejection alone may leave no outcome signal. The bottom panel of Figure~\ref{fig:overview} illustrates these contrasting learning dynamics. We therefore evaluate reward accuracy together with correct-reward retention, active training groups, and model performance.

\section{Method}
\label{sec:method}
\subsection{Problem Setup and Overview}
We assume a labeled source set $\mathcal{D}_s=\{(x,a^*)\}$ and a target set $\mathcal{D}_t$ whose ground-truth answers are unavailable for reward computation. Rectifier fitting and subsequent test-time adaptation use disjoint sets of questions. TRUER first trains a rectifier $V_\phi$ on source candidates, then freezes $\phi$ while adapting a language model policy $\pi_\theta$ on target inputs (Figure~\ref{fig:overview}).

\subsection{Learning the Reward Rectifier}
Each rectifier example contains a question, a candidate solution, and its extracted answer. Source ground truth determines correctness, while the sampled candidate pool determines majority membership. Their cross-product gives four categories: majority-correct, majority-wrong, minority-correct, and minority-wrong. We emphasize the two categories where consensus and correctness disagree, while retaining examples of correct majorities and incorrect minorities.

The rectifier produces a short justification and an integer score $s\in\{0,\ldots,9\}$. With $p=s/9$ and correctness label $c\in\{0,1\}$, its training reward is
\begin{equation}
 R_V=1-2(p-c)^2-0.1\,\ind[\text{explicit reason--score conflict}].
 \label{eq:verifier}
\end{equation}
Malformed or truncated outputs receive a fixed penalty. GRPO optimizes the rectifier using multiple judgments per candidate. Category weights emphasize incorrect majorities and correct minorities; they are applied to the policy loss after within-group advantage normalization. Neither the gold answer nor the category label is included in the rectifier input. The weighting formula and recorded training settings are given in Appendix~\ref{app:rectifiertraining}.

\subsection{Rectified Rewards for Target Adaptation}
The model samples responses and groups equivalent final answers. TRUER checks an eligible representative of the largest cluster, rank1. A valid score $s\geq\tau$ accepts that answer, with threshold $\tau=5$ in our experiments. An explicit rejection ($s<\tau$) triggers a rank2 check; an unresolved judgment does not. If neither branch selects an answer, TRUER abstains. Ties and candidate eligibility are handled by fixed rules described in Appendix~\ref{app:protocol}.

For selected answer $\hat a$, with $\bot$ denoting abstention, the reward and outcome advantage are
\begin{equation}
 r_i^{\mathrm{TRUER}}=\ind[\hat a\neq\bot]\ind[a_i\equiv\hat a],\qquad
 A_i=\frac{r_i^{\mathrm{TRUER}}-\overline r^{\,\mathrm{TRUER}}}
 {\operatorname{std}(r^{\mathrm{TRUER}})+10^{-6}}.
 \label{eq:advantage}
\end{equation}
The policy is updated with GRPO while the rectifier remains frozen (Table~\ref{alg:truer}). Abstention assigns all-zero rewards and removes the group's direct outcome advantage.

\begin{table}[t]
\centering
\caption{\textbf{TRUER target adaptation.} The rectifier is trained on source labels before this procedure.}
\label{alg:truer}
\begin{minipage}{.94\linewidth}
\hrule\smallskip
\textbf{Input:} policy $\pi_\theta$, frozen rectifier $V_\phi$, target questions $\mathcal{D}_t$.\\
\textbf{For each policy update:}
\begin{enumerate}
\item Sample a batch of target questions and candidate solutions from $\pi_\theta$.
\item Group equivalent answers and rank the clusters by frequency.
\item Initialize $\hat a=\bot$. Verify rank1 when its route is eligible and untied.
\item If rank1 is accepted, select it. If explicitly rejected, verify an eligible, untied rank2 and select it only if accepted.
\item Assign rewards using Equation~\ref{eq:advantage} and update $\theta$ with GRPO.
\end{enumerate}
\textbf{Output:} adapted policy and the unchanged rectifier.
\smallskip\hrule
\end{minipage}
\end{table}

\section{Experiments}
\label{sec:experiments}
% Native 9pt table: no whole-table scaling. Data: assets/main_results/data/main_table.json.
\definecolor{resultblue}{HTML}{126CB2}
\definecolor{resultred}{HTML}{C8322E}
\definecolor{resultgray}{HTML}{707070}
\newcommand{\passgap}[4]{\begin{tikzpicture}[x=1cm,y=1cm,baseline=-.4ex]
\pgfmathsetmacro{\gapscale}{max(1,#1)}
\pgfmathsetmacro{\oraclex}{2.8/\gapscale}
\pgfmathsetmacro{\oursx}{2.8*#1/\gapscale}
\draw[black!55] (0,0)--(\oraclex,0);
\draw[black!55] (0,-.025)--(0,.025) (\oraclex,-.025)--(\oraclex,.025);
\draw[resultblue,line width=.9pt] (0,0)--(\oursx,0);
\fill[resultblue] (\oursx,0) circle[radius=1.6pt];
\node[above,font=\fontsize{7.5}{8}\selectfont,text=resultred,inner sep=1.5pt] at (1.4,.02){+#2 pp};
\node[below,font=\fontsize{7}{8}\selectfont,inner sep=1.5pt] at (0,-.025){TTRL};
\node[below,font=\fontsize{7}{8}\selectfont,inner sep=1.5pt] at (\oraclex,-.025){Oracle};
\node[below,font=\fontsize{7}{8}\selectfont,text=resultred,inner sep=1.5pt] at (1.4,-.025){#3 pp / #4\%};
\end{tikzpicture}}
\begin{table*}[!t]
\centering
\caption{\textbf{TRUER main results (\%).} Columns report the model/task, rectifier, Pass@1, Pass@32, and reward accuracy. In each Pass@1 bar, the gray line spans the TTRL--Oracle gap and the blue line shows the TTRL--TRUER gain. Labels give the gain in percentage points (above), and the full gap in percentage points and the fraction closed (below).}
\label{tab:core-results}
\begingroup
\fontsize{9}{10.5}\selectfont
\setlength{\tabcolsep}{1.6pt}
\renewcommand{\arraystretch}{1.10}
\begin{tabular}{@{}l>{\raggedright\arraybackslash}p{52pt}rrrrrrrrr@{}}
\toprule
\textbf{Model / task} & \textbf{Rectifier} & \multicolumn{4}{c}{\textbf{Pass@1}} & \multicolumn{3}{c}{\textbf{Pass@32}} & \multicolumn{2}{c}{\textbf{Reward acc.}} \\
\cmidrule(lr){3-6}\cmidrule(lr){7-9}\cmidrule(lr){10-11}
 & & Base & TTRL & TRUER & Oracle & TTRL & TRUER & Oracle & TTRL & TRUER \\
\midrule
\multirow{2}{*}{\shortstack[l]{Qwen2.5-Math\\1.5B / MATH500}} & \multirow{2}{*}{Qwen3-4B} & 21.85 & 69.62 & \textcolor{resultblue}{\textbf{70.46}} & \textcolor{resultgray}{70.87} & 92.97 & \textcolor{resultblue}{\textbf{91.89}} & \textcolor{resultgray}{93.24} & 95.53 & \textcolor{resultblue}{\textbf{96.36}} \\
 & & \multicolumn{4}{c}{\passgap{0.6756756756756719}{0.84}{1.25}{67.6}} & & & & & \\
\addlinespace[1.5pt]\midrule
\multirow{2}{*}{\shortstack[l]{Qwen3-1.7B\\AMC23}} & \multirow{2}{*}{Qwen3-1.7B} & 46.09 & 56.09 & \textcolor{resultblue}{\textbf{57.11}} & \textcolor{resultgray}{64.61} & 77.50 & \textcolor{resultblue}{\textbf{92.50}} & \textcolor{resultgray}{92.50} & 74.84 & \textcolor{resultblue}{\textbf{86.09}} \\
 & & \multicolumn{4}{c}{\passgap{0.11926605504587123}{1.02}{8.52}{11.9}} & & & & & \\
\addlinespace[1.5pt]\midrule
\multirow{2}{*}{\shortstack[l]{Qwen3-1.7B\\AMC23}} & \multirow{2}{*}{Qwen3-4B} & 46.09 & 56.09 & \textcolor{resultblue}{\textbf{59.38}} & \textcolor{resultgray}{64.61} & 77.50 & \textcolor{resultblue}{\textbf{92.50}} & \textcolor{resultgray}{92.50} & 74.84 & \textcolor{resultblue}{\textbf{89.22}} \\
 & & \multicolumn{4}{c}{\passgap{0.38532110091743127}{3.28}{8.52}{38.5}} & & & & & \\
\addlinespace[1.5pt]\midrule
\multirow{2}{*}{\shortstack[l]{Qwen3-1.7B\\AIME24}} & \multirow{2}{*}{Qwen3-1.7B} & 13.96 & 20.21 & \textcolor{resultblue}{\textbf{21.46}} & \textcolor{resultgray}{24.69} & 46.67 & \textcolor{resultblue}{\textbf{50.00}} & \textcolor{resultgray}{46.67} & 61.41 & \textcolor{resultblue}{\textbf{86.80}} \\
 & & \multicolumn{4}{c}{\passgap{0.2790697674418601}{1.25}{4.48}{27.9}} & & & & & \\
\addlinespace[1.5pt]\midrule
\multirow{2}{*}{\shortstack[l]{Qwen3-1.7B\\AIME24}} & \multirow{2}{*}{Qwen3-4B} & 13.96 & 20.21 & \textcolor{resultblue}{\textbf{24.79}} & \textcolor{resultgray}{24.69} & 46.67 & \textcolor{resultblue}{\textbf{43.33}} & \textcolor{resultgray}{46.67} & 61.41 & \textcolor{resultblue}{\textbf{88.28}} \\
 & & \multicolumn{4}{c}{\passgap{1.0232558139534882}{4.58}{4.48}{102.3}} & & & & & \\
\addlinespace[1.5pt]\midrule
\multirow{2}{*}{\shortstack[l]{Qwen3-4B\\AMC23}} & \multirow{2}{*}{Qwen3-4B} & 65.39 & 79.69 & \textcolor{resultblue}{\textbf{82.34}} & \textcolor{resultgray}{85.63} & 92.50 & \textcolor{resultblue}{\textbf{97.50}} & \textcolor{resultgray}{97.50} & 87.73 & \textcolor{resultblue}{\textbf{97.66}} \\
 & & \multicolumn{4}{c}{\passgap{0.44736842105263264}{2.66}{5.94}{44.7}} & & & & & \\
\addlinespace[1.5pt]\midrule
\multirow{2}{*}{\shortstack[l]{Qwen3-4B\\MATH-L5}} & \multirow{2}{*}{Qwen3-4B} & 69.43 & 79.82 & \textcolor{resultblue}{\textbf{80.68}} & \textcolor{resultgray}{83.76} & 92.50 & \textcolor{resultblue}{\textbf{93.00}} & \textcolor{resultgray}{94.83} & 87.81 & \textcolor{resultblue}{\textbf{89.92}} \\
 & & \multicolumn{4}{c}{\passgap{0.21825396825396687}{0.86}{3.94}{21.8}} & & & & & \\
\addlinespace[1.5pt]\midrule
\multirow{2}{*}{\shortstack[l]{Llama-3.1-8B\\Instruct / MATH500}} & \multirow{2}{*}{Qwen3-4B} & 37.02 & 63.93 & \textcolor{resultblue}{\textbf{73.61}} & \textcolor{resultgray}{83.68} & 67.03 & \textcolor{resultblue}{\textbf{85.95}} & \textcolor{resultgray}{91.62} & 75.66 & \textcolor{resultblue}{\textbf{96.12}} \\
 & & \multicolumn{4}{c}{\passgap{0.49038050448909765}{9.69}{19.76}{49.0}} & & & & & \\
\addlinespace[1.5pt]\midrule
\multirow{2}{*}{\shortstack[l]{Llama-3.1-8B\\Instruct / AMC23}} & \multirow{2}{*}{Qwen3-4B$^{\dagger}$} & 12.42 & 49.45 & \textcolor{resultblue}{\textbf{60.23}} & \textcolor{resultgray}{88.20} & 50.00 & \textcolor{resultblue}{\textbf{75.00}} & \textcolor{resultgray}{95.00} & 61.48 & \textcolor{resultblue}{\textbf{89.48}} \\
 & & \multicolumn{4}{c}{\passgap{0.2782258064516128}{10.78}{38.75}{27.8}} & & & & & \\
\bottomrule
\end{tabular}
\endgroup
\par\vspace{3pt}\noindent
\begin{minipage}{\textwidth}\footnotesize
$^{\dagger}$Same rectifier as the Llama/MATH500 configuration above.
\end{minipage}
\end{table*}

\subsection{Experimental Setup}
\paragraph{Models and tasks.}
We evaluate Qwen2.5-Math-1.5B and Qwen3 models with 1.7B or 4B parameters \citep{yang2024qwenmath,qwen2025qwen3}, as well as Llama-3.1-8B-Instruct \citep{grattafiori2024llama3}, using Qwen3 rectifiers with 1.7B or 4B parameters. Table~\ref{tab:core-results} contains nine configurations across MATH500 and MATH-L5, both drawn from MATH \citep{hendrycks2021math}, as well as AMC23 and AIME24. The MATH500 experiments reserve 130 questions for rectifier development and use the remaining 370 for target adaptation and evaluation. MATH-L5 uses a separate set of 600 questions. For Qwen models on AMC23 and AIME24, rectifiers use source questions from AMC2024 and AIME2025, respectively. Llama on AMC23 reuses a frozen rectifier trained solely on MATH. Detailed experimental protocols are provided in Appendix~\ref{app:mainprotocol}.

\paragraph{Comparisons and metrics.}
\textbf{Naive TTRL} implements the majority-vote reward mechanism of \citet{zuo2025ttrl} under our experimental protocol; \textbf{TRUER} uses rectified rewards; \textbf{Oracle} is a supervised reference using target ground-truth rewards with GRPO \citep{shao2024deepseekmath} under the corresponding policy training configuration. Base is the model before adaptation. Pass@1 estimates single-sample accuracy from the saved responses, Pass@32 measures candidate coverage, and Majority@32 measures voting accuracy on the same pool \citep{chen2021code}. Reward accuracy is the fraction of binary rewards agreeing with response correctness, aggregated over the training windows specified in Appendix~\ref{app:mainprotocol}. The main table reports each model's final checkpoint; the mechanism audit instead compares reward policies on one fixed response pool.

\paragraph{Evaluation scope.}
All reported scores are measured in our own experiments, including the Base, naive TTRL, and Oracle references. We evaluate model performance on the target adaptation sets. Target labels are excluded from TRUER's reward decisions.

\subsection{Main Results}
\label{sec:oracle}
Across all nine configurations in Table~\ref{tab:core-results}, TRUER improves reward accuracy and final Pass@1 over naive TTRL, narrowing the Pass@1 gap to the Oracle reference trained with ground-truth rewards. For example, using a Qwen3-4B rectifier, Qwen2.5-Math-1.5B on MATH500 improves reward accuracy from 95.53\% to 96.36\% and closes 67.6\% of the TTRL--Oracle gap; Qwen3-1.7B on AMC23 improves reward accuracy from 74.84\% to 89.22\% and closes 38.5\% of the gap. The final Llama-3.1-8B-Instruct/AMC23 row reuses a MATH-trained rectifier without AMC-specific rectifier training, suggesting some transferability across datasets.

Pass@32 varies less consistently. Where it changes little despite higher Pass@1, reinforcement learning may sharpen the answer distribution toward correct solutions already reachable by the base model, improving sampling efficiency without necessarily expanding its reasoning coverage. This interpretation is consistent with prior observations on RLVR \citep{yue2025reasoningcapacity}.

\subsection{Case Study: Reward Correction and Model Adaptation}
\label{sec:fixedpool}
\label{sec:mechanism}
We investigate Qwen2.5-Math-1.5B with a Qwen3-4B rectifier on MATH500 to examine how reward correction relates to model adaptation.

\begin{figure}[t]
\centering
\includegraphics[width=\linewidth]{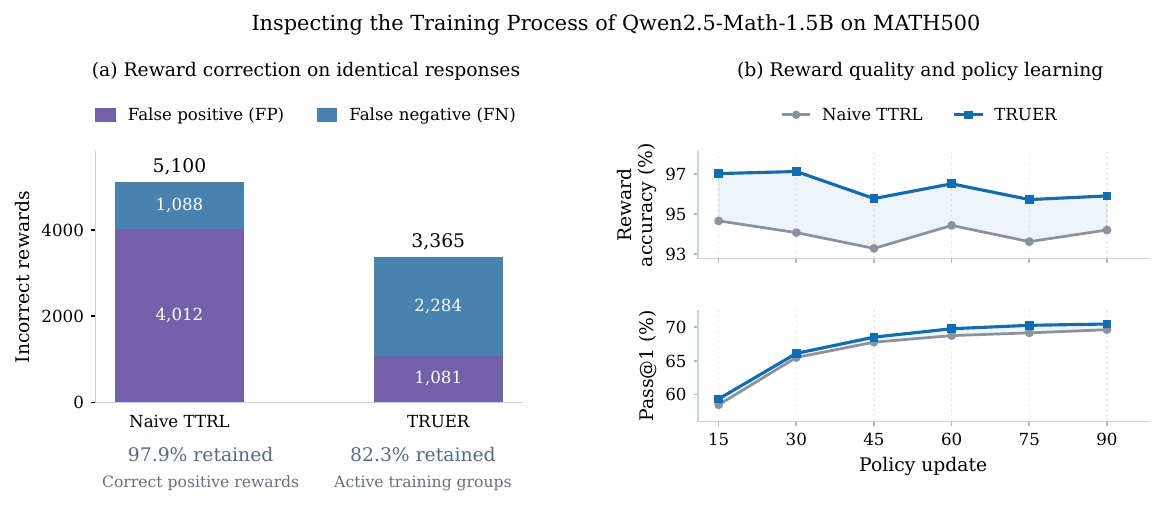}
\caption{\textbf{Reward correction on MATH500.} Qwen2.5-Math-1.5B with a Qwen3-4B rectifier. (a) FP/FN counts on identical responses; retention is relative to majority voting. (b) Reward accuracy over 15-update windows and endpoint Pass@1 on each method's own trajectory.}
\label{fig:rewardlearning}
\label{fig:rewardaudit}
\label{fig:training}
\end{figure}

\paragraph{Reward accuracy and model performance.}
We first apply naive majority rewards and TRUER to the same 92,160 responses from the TRUER training run. As Figure~\ref{fig:rewardlearning}(a) shows, total reward errors fall from 5,100 to 3,365, with 73.1\% fewer false positives, although false negatives increase. TRUER retains 97.9\% of the correct-positive reward count, including correct rewards recovered through rank2, and 82.3\% of active training groups relative to majority voting. Vote-margin filters retain fewer correct rewards at comparable false-positive removal (Appendices~\ref{app:reward} and~\ref{app:margin}).

Across training, TRUER achieves higher reward accuracy in every 15-update window and higher Pass@1 at all six checkpoints (Figure~\ref{fig:rewardlearning}(b)). Appendix~\ref{app:results} provides additional learning curves and uncertainty estimates.

\begin{figure}[t]
\centering
\includegraphics[width=\linewidth]{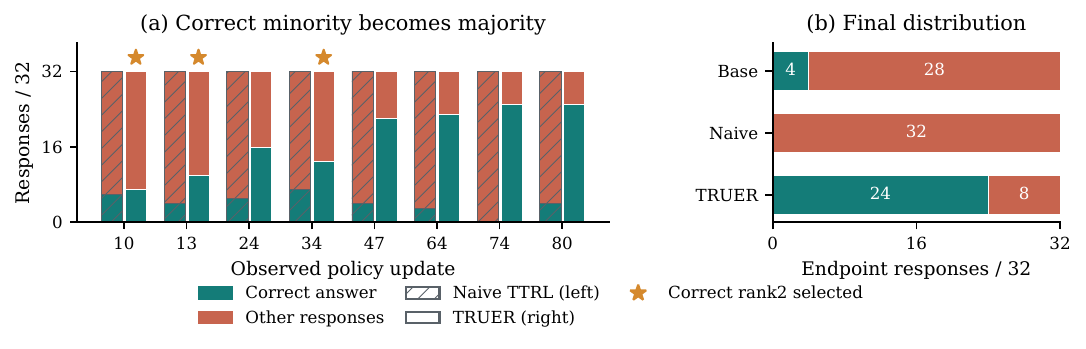}
\caption{\textbf{From correct minority to majority.} (a) Each training visit pairs naive TTRL (left, hatched) with TRUER (right, solid); stars mark correct rank2 selection. (b) Separately sampled endpoint responses. Each bar contains 32 responses; ``Other responses'' includes incorrect and unparsed answers.}
\label{fig:cases}
\end{figure}

\paragraph{A concrete case: correct minority becomes majority.}
Figure~\ref{fig:cases} tracks both methods on the same question at matched training steps. TRUER rejects an incorrect majority and rewards the correct minority, which later becomes dominant, while naive TTRL retains the wrong consensus. Recovery is not guaranteed: repeated abstention can leave the model incorrect. Appendix~\ref{app:cases} gives both examples and their selection rules, while Appendix~\ref{app:distribution} reports aggregate answer-distribution changes.

\subsection{Additional Evaluation: Rectifier Reuse at Deployment}
\label{sec:inference}
\begin{figure}[t]
\centering
\includegraphics[width=\linewidth]{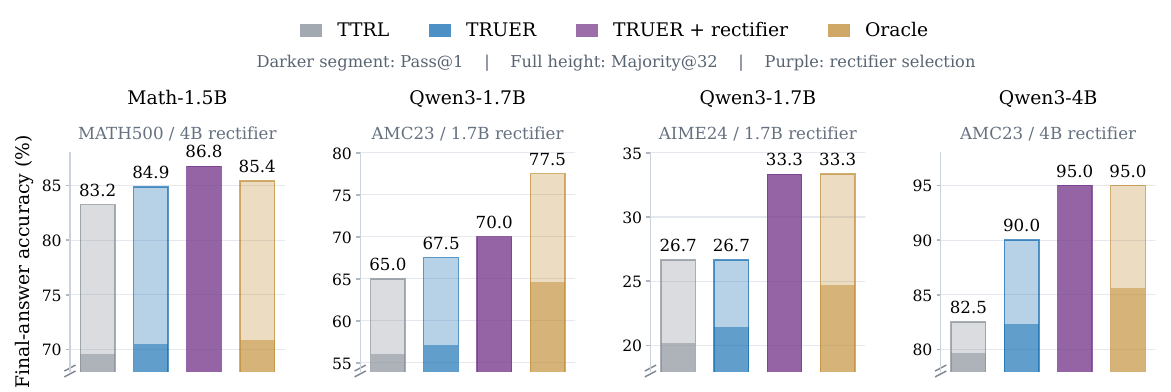}
\caption{\textbf{Rectifier reuse at deployment.} For TTRL, TRUER, and Oracle, darker segments end at Pass@1 and lighter extensions reach Majority@32. Purple bars show TRUER with score-weighted answer aggregation on identical 32-candidate pools. Panels use different truncated y-axis ranges. Four configurations are shown; Appendix~\ref{app:inference} reports all six.}
\label{fig:deployment}
\end{figure}

Scoring candidate solutions is a common use of test-time computation \citep{cobbe2021verifiers,snell2024scaling}. We therefore evaluate whether the rectifier trained for reward correction can also support answer selection at deployment. Given 32 sampled candidates, we group equivalent answers and sum their thresholded rectifier scores, normalized by the square root of the number of valid judgments in each group. We select the answer with the largest aggregate support, with majority fallback if no candidate passes. Figure~\ref{fig:deployment} compares this strategy with single-sample answering and majority voting for naive TTRL, TRUER, and Oracle. Under this shared rule, selected through exploratory target-set replay, rectifier selection improves over TRUER majority voting in all six evaluated configurations, by 0.17--6.67 percentage points. With 32 candidates, it exceeds Oracle single-sample accuracy in all six configurations. Appendix~\ref{app:inference} specifies the aggregation rule, its selection procedure, matched naive-TTRL results, and inference cost.

\subsection{Ablations and Limitations}
\label{sec:rank1}
\paragraph{Effect of rectifier training.}
In our fixed-pool comparisons, the trained rectifier achieves higher judgment accuracy and rejects more incorrect answers than its pretrained counterpart. For example, on responses generated by naive TTRL, training increases judgment accuracy from 71.85\% to 86.52\% and reduces wrong-answer acceptance from 52.11\% to 18.35\%. Detailed results and evaluation settings are provided in Appendix~\ref{app:verifiertraining}. For adaptation, TRUER with a trained rectifier achieves higher final Pass@1 than its untrained-rectifier variant in the three comparisons reported in Table~\ref{tab:untrained-adaptation}. On AIME24 with the 1.7B rectifier, the untrained variant even falls below naive TTRL (18.96\% vs.\ 20.21\%), whereas the trained variant reaches 21.46\%.

\paragraph{Training overhead.}
Although deploying the adapted model alone incurs no additional rectifier computation, reward correction increases training time. For Qwen2.5-Math-1.5B on MATH500, the recorded mean time per policy update increases from 177.2 seconds for naive TTRL to 219.3 seconds for TRUER, an overhead of approximately 23.7\% with a dedicated rectifier GPU, excluding rectifier pretraining (Appendix~\ref{app:budgetcontrols}).

\paragraph{Training-budget controls.}
We extend naive TTRL to approximately match TRUER's adaptation-stage GPU-hours, and separately extend TRUER to match the original naive TTRL run's cumulative number of groups with nonzero outcome advantages, to assess TRUER's performance relative to naive TTRL under comparable computation and cumulative counts of advantage-bearing groups, respectively. Additional naive training narrows the gap in some settings, but TRUER retains higher final Pass@1 in all five GPU-hour-matched comparisons. The active-group-matched controls likewise preserve TRUER's advantage over the original naive TTRL checkpoints. Full results and protocols are provided in Appendix~\ref{app:budgetcontrols}.

\paragraph{Evaluation scope.}
Our evaluation focuses on mathematical reasoning; extending reward rectification to open-ended tasks with less structured answers remains an interesting direction.

\section{Conclusion}
We introduced TRUER, which uses limited source supervision to correct unreliable consensus rewards in test-time reinforcement learning without access to target ground-truth labels. By emphasizing incorrect majorities and correct minorities during rectifier training, our method improves reward accuracy, which in turn leads to better model performance. Across all evaluated configurations, TRUER improves model performance over naive TTRL and narrows the gap to the Oracle reference trained with ground-truth rewards. This offers a practical path toward adaptation where ground-truth feedback for new problems is costly or infeasible.

\clearpage
\subsection*{AI use statement}
We used generative AI tools to assist with experimental design, code implementation, and selected aspects of result analysis, as well as manuscript formatting checks, verification of cross-references and bibliographic metadata, and editorial revisions. The authors take full responsibility for the experimental design, code, reported results and their interpretation, and the final manuscript.

\bibliography{iclr2027_conference}
\bibliographystyle{iclr2027_conference}
\clearpage
\appendix
\label{document:appendix-start}
% Keep natural spacing around appendix text and in-place figures/tables.
% Otherwise flush-bottom justification stretches sparse pages between blocks.
\raggedbottom

\section{Experimental Setup and Protocols}
\label{app:mainprotocol}
\label{app:data}
\label{app:amc}
\subsection{Configurations and Supervision}
Table~\ref{tab:protocolsummary} records the nine main-table configurations. The six matched deployment comparisons use the same score-weighted answer aggregation rule with 32 candidates and majority fallback (denoted Weighted in Table~\ref{tab:protocolsummary}). Source-label budgets vary across rows.

\begin{table}[H]
\centering\small
\caption{\textbf{Configurations in the nine-row main table.} Seeds describe policy training. Rectifier steps identify the frozen checkpoint, not additional policy updates. A dash denotes a configuration outside the matched deployment comparison.}
\label{tab:protocolsummary}
\setlength{\tabcolsep}{4pt}
\begin{tabular}{llrrlrl}
\toprule
Model & Target & Seed & Updates & Rectifier & Step & Deploy \\
\midrule
Math-1.5B & MATH500 & 42 & 90 & Qwen3-4B & 150 & Weighted \\
Qwen3-1.7B & AMC23 & 42 & 120 & Qwen3-1.7B & 198 & Weighted \\
Qwen3-1.7B & AMC23 & 42 & 120 & Qwen3-4B & 25 & Weighted \\
Qwen3-1.7B & AIME24 & 42 & 120 & Qwen3-1.7B & 260 & Weighted \\
Qwen3-1.7B & AIME24 & 42 & 120 & Qwen3-4B & 312 & --- \\
Qwen3-4B & AMC23 & 43 & 120 & Qwen3-4B & 175 & Weighted \\
Qwen3-4B & MATH-L5 & 43 & 240 & Qwen3-4B & 50 & Weighted \\
Llama-3.1-8B & MATH500 & 43 & 120 & Qwen3-4B & 175 & --- \\
Llama-3.1-8B & AMC23 & 43 & 120 & Qwen3-4B & 175 & --- \\
\bottomrule
\end{tabular}
\end{table}

\paragraph{Qwen2.5-Math-1.5B on MATH500.}
We reserve 130 MATH500 questions for rectifier development and use the remaining 370 for test-time adaptation and evaluation. The eligible source pool contains 3,648 candidates from 124 questions. The selected rectifier consumes 2,400 distinct candidates and generates 21,016 judgment trajectories through update 150. Target questions are excluded from rectifier fitting by ID, normalized text, and the recorded question-family check.

Source candidates were generated by a model after one warm-up round on all MATH500 inputs, without access to any ground-truth answers. Rectifier training and checkpoint selection were then performed exclusively on source data. The models used for evaluation restart from pretrained weights.

\paragraph{AMC23.}
The configurations using the 1.7B model fit rectifiers on 528 candidates from 20 AMC2024 questions, with ten further AMC2024 questions for source calibration. Test-time adaptation and evaluation use the same 40 AMC23 questions. The source and target files have no overlapping IDs, normalized texts, or digit-template families. The 4B rectifier is frozen at step 25. The main adaptation run with the 4B rectifier uses seed 42; its naive TTRL and Oracle references are shared with the 1.7B-rectifier configuration.

The stable 1.7B rectifier uses the same source data and category weights, learning rate $10^{-6}$ with ten-step warmup, and KL coefficient 0.02. Its step 198 checkpoint is frozen for test-time adaptation. The model uses seed 42, with historical Naive and Oracle controls; its random stream differs from those controls. This is not an isolated rectifier-size comparison.

\paragraph{Training repetitions.}
For the Qwen3-1.7B model with the 4B rectifier on AMC23, naive TTRL, TRUER, and Oracle were each independently trained from the same pretrained policy weights for each seed (42, 43, and 44), yielding nine runs evaluated at the final update 120. The rectifier was held fixed across seeds. Table~\ref{tab:amc-training-seeds} reports all three methods for each seed. TRUER improves Pass@1 over the independently trained naive TTRL run with the same seed by 3.28, 5.78, and 4.84 percentage points, respectively (4.64 points on average). The main table reports seed 42 for consistency with the historical naive TTRL and Oracle controls; this is the most conservative of the three runs in terms of Pass@1 gain.

\begin{table}[H]
\centering\small
\caption{AMC23 training repetitions: Qwen3-1.7B with the fixed 4B rectifier. Pass@1 (\%) at update 120. Each row contains independently trained policies using the same training seed. $\Delta$ is TRUER minus naive TTRL in percentage points, computed before rounding.}
\label{tab:amc-training-seeds}
\begin{tabular}{lrrrr}
\toprule
Seed & Naive TTRL & TRUER & Oracle & $\Delta$ \\
\midrule
42 & 56.09 & 59.38 & 64.61 & +3.28 \\
43 & 55.23 & 61.02 & 64.30 & +5.78 \\
44 & 54.84 & 59.69 & 65.31 & +4.84 \\
Mean & 55.39 & 60.03 & 64.74 & +4.64 \\
\bottomrule
\end{tabular}
\end{table}

\paragraph{AIME24.}
Both AIME24 configurations use a Qwen3-1.7B model, 30 AIME2025 source questions for rectifier fitting, and all 30 AIME2024 questions for test-time adaptation and evaluation. The 1.7B and 4B rectifiers are frozen at epoch 5 (step 260) and epoch 6 (step 312), respectively. The two configurations share the same Base, naive TTRL, and Oracle references. All policy training runs use seed 42 and 40 epochs (120 updates), with ten questions per update and 32 responses per question. Response generation uses temperature 1, a 2,048-token prompt limit, and a 4,096-token response limit. Main-table reward accuracy uses updates 117--120: out of 1,280 assignments per arm, Naive has 786 correct, and TRUER has 1,111 with the 1.7B rectifier and 1,130 with the 4B rectifier. The latter model's Pass@1 is 238/960, compared with 237/960 for Oracle, yielding 102.3\% gap closure; Oracle is an empirical reference rather than a strict performance ceiling.

\paragraph{Additional 4B configurations.}
The 4B AMC row uses the recorded mixed-100 source rectifier. MATH-L5 uses a separate 600-question target set. The retained summary does not itemize the complete source-label budgets for these two rectifiers; neither inherits the MATH130 or AMC20 budget by assumption.

\paragraph{Llama-3.1-8B-Instruct on MATH500.}
The Llama configuration uses the fixed MATH500 split of 100 source-fitting questions, 30 source-calibration questions, and 370 target questions. We generate 64 source candidates per question with Llama before fitting the Qwen3-4B rectifier. Rectifier step 175 is selected using source calibration and frozen for test-time adaptation. All three adaptation runs use seed 43, 120 updates over ten epochs, batches of up to 32 questions with 32 responses per question, learning rate $5\times10^{-7}$, three warmup steps, KL coefficient 0.001, a 4,096-token response limit, and full-parameter training.

All endpoint outputs are scored with the same retained scorer. Evaluation uses four logical shards, eight-question batches, and question-specific seeds. Deployment verification has not been measured.

\paragraph{Llama-3.1-8B-Instruct on AMC23.}
This configuration reuses the frozen Qwen3-4B step 175 rectifier from the Llama/MATH500 experiment without AMC-specific retraining. Rectifier fitting uses 100 MATH500 source questions, with 64 candidates per question generated by the original Llama model and labeled using source ground truth. All three AMC adaptation runs start from the original Llama-3.1-8B-Instruct model, use seed 43, and complete 120 updates on eight GPUs. Adaptation and endpoint evaluation use all 40 AMC23 questions, with 32 responses per question for the sampled evaluation. These are model endpoint scores; deployment rectifier selection has not been measured for this configuration.

\paragraph{Experimental coverage.}
Hardware availability and time constraints limited our coverage of the full matrix of language models, rectifiers, tasks, and training seeds. Nevertheless, TRUER achieves higher model Pass@1 and reward accuracy than naive TTRL in all nine configurations reported in Table~\ref{tab:core-results}. We view these results, together with the three-seed AMC23 comparison, as supporting the effectiveness of the method within the evaluated settings.

\subsection{Metrics and Recorded Protocols}
For $c_q$ correct answers among $N=32$ responses to question $q$,
\begin{align}
\widehat{\mathrm{Pass@1}}&=\frac{1}{Q}\sum_q\frac{c_q}{32},&
\widehat{\mathrm{Pass@32}}&=\frac{1}{Q}\sum_q\ind[c_q>0],\\
\widehat{\mathrm{Pass@2}}&=\frac{1}{Q}\sum_q\left[1-\frac{(32-c_q)(31-c_q)}{32\cdot31}\right].
\end{align}
Majority@32 uses equivalent-answer clusters from the same responses. Unparsed responses remain in the denominator. The main table uses the final model checkpoint, with no deployment verification in the Pass metrics. Gap closure is $(\mathrm{Pass@1}_{\mathrm{TRUER}}-\mathrm{Pass@1}_{\mathrm{Naive}})/(\mathrm{Pass@1}_{\mathrm{Oracle}}-\mathrm{Pass@1}_{\mathrm{Naive}})$, computed before rounding.

Main-table reward accuracy covers each Qwen arm's final four updates: 87--90 for Qwen2.5-Math-1.5B/MATH500, 117--120 for AMC23 and AIME24, and 237--240 for L5. These contain 4,096 responses per Qwen2.5-Math arm and 1,280 per other Qwen arm, including constant-reward groups. The Llama/MATH500 row uses the complete 120-update audit, with 118,400 responses per arm: naive TTRL has 69,295 TP, 27,737 FP, 20,284 TN, and 1,084 FN; TRUER has 75,309 TP, 2,208 FP, 38,500 TN, and 2,383 FN. Reward accuracy is 75.66\% and 96.12\%, respectively. The Llama/AMC23 row also uses all 120 updates, with 38,400 responses per arm: naive TTRL has 15,710 TP, 14,571 FP, 7,898 TN, and 221 FN; TRUER has 15,215 TP, 1,085 FP, 19,144 TN, and 2,956 FN, giving reward accuracies of 61.48\% and 89.48\%, respectively. These counts come from each arm's own trajectory; fixed-pool audits instead compare labeling rules on identical saved responses.

The Qwen2.5-Math-1.5B settings appear in Table~\ref{tab:trainingdetails}. AMC adaptation runs use 120 updates, ten questions per update, 32 responses per question, a 4,096-token response cap, and a 5,120-token context. The rectifier uses greedy decoding with a 4,096-token output cap and 8,192-token context. The L5 endpoint is update 240. Reward selection never receives target gold; Oracle uses it explicitly.

\begin{table}[H]
\centering
\caption{Qwen2.5-Math-1.5B/MATH500 training settings.}\label{tab:trainingdetails}
\begin{tabular}{p{.35\linewidth}p{.55\linewidth}}
\toprule
Setting & Value \\
\midrule
Model & Qwen2.5-Math-1.5B, original pretrained weights \\
Policy trainable parameters & All parameters; four-GPU FSDP2 \\
Policy optimizer & AdamW; weight decay 0.01; gradient clipping 1.0 \\
Policy learning rate & Peak $5\times10^{-7}$; three warmup steps; cosine to step 90 \\
Reference policy / KL & Frozen initial policy; separate KL loss coefficient 0.001 \\
Training batch / responses & 32 questions / 32 responses per question \\
Generation settings & Temperature 1; top-$p$ 1; top-$k$ disabled \\
Generation lengths & Prompt 1,024; response 3,072; context 4,096 tokens \\
Rectifier & Qwen3-4B; LoRA \citep{hu2021lora} rank 64, scaling alpha 128 \\
Rectifier optimizer & Learning rate $5\times10^{-6}$; weight decay 0.01 \\
Rectifier training groups & 16 candidate prompts per batch; initially 8 judgments per candidate \\
Rectifier extra sampling & Same-policy expansion to 16/32 for homogeneous-reward groups that are not all successful \\
Rectifier PPO / KL & Clip 0.2; KL coefficient 0.01 \\
Rectifier decoding & Training temperature 1; native reasoning; output limit 4,096, context 8,192 \\
Rectifier deployment & Frozen step 150; greedy reasoning; score $\geq5$ accepts \\
Evaluation & Sampled $N=32$, temperature 1; greedy separately at temperature 0 \\
Evaluation scheduling & Four-question batches; fixed question-specific hash seeds \\
\bottomrule
\end{tabular}
\end{table}

MATH TTRL and TRUER resume at step 30 with optimizer, scheduler, and random states restored; rank1-only and Oracle run continuously. TTRL resolves top-count ties by sample order, whereas rectifier policies abstain. The rank1-only/TRUER ablation shares the tie rule.

\paragraph{Scoring and uncertainty.}
We retain the frozen mathematical scorer across arms. Known edge cases involving intervals, units, equations, and numeric bases are not silently corrected in the main results. An audit affected sixteen rectifier candidate labels, including four training candidates; the selected MATH rectifier was not retrained after this audit. Reviewed examples exclude known scoring errors. Paired bootstrap intervals resample question IDs 10,000 times and do not include training-seed or checkpoint-selection uncertainty.

\subsection{Rectifier Training and Selection Rules}

\label{app:protocol}
\label{app:rectifiertraining}
The rectifier receives the problem, candidate solution, and extracted answer. Its prompt asks for an independent check and a final JSON judgment containing a reason and score. It neither executes candidate code nor receives gold answers or category labels. Scores are learned ratings rather than calibrated probabilities. Valid outputs use Equation~\ref{eq:verifier}; malformed or truncated outputs receive $-1.2$.

For the rectifier used with Qwen2.5-Math-1.5B on MATH500, category $k$ with $n_k$ candidates and priority $\alpha_k$ has weight
\begin{equation}
 w_k=\frac{M\alpha_k}{n_k\sum_j\alpha_j},\qquad M=3{,}648.
\end{equation}
The weight multiplies the policy loss after advantage normalization. Table~\ref{tab:quadrants} describes the eligible candidate pool.

\begin{table}[H]
\centering
\caption{\textbf{Rectifier supervision for Qwen2.5-Math-1.5B/MATH500.} Correctness labels come from source ground truth; category membership is not provided as input.}
\label{tab:quadrants}
\begin{tabular}{lrrr}
\toprule
Category & Eligible candidates & Priority & Loss weight \\
\midrule
Majority-correct & 1,621 & 1 & 0.2813 \\
Majority-wrong & 440 & 3 & 3.1091 \\
Minority-correct & 153 & 3 & 8.9412 \\
Minority-wrong & 1,434 & 1 & 0.3180 \\
\bottomrule
\end{tabular}
\end{table}

During test-time adaptation, rank1 is checked using its first eligible representative in sampling order. Eligibility uses parsing and length. A valid score $s\geq5$ accepts the answer; $s<5$ permits a rank2 check if there is no boundary tie. Missing representatives, unresolved judgments, and rejected alternatives yield abstention. Rank1-only removes the second check. Abstained groups have zero outcome advantage; the separate KL term remains active.

\section{Additional Results and Learning Dynamics}
\label{app:results}

Original figure legends labeled ``Ours'' or ``Full cascade'' refer to TRUER.

\subsection{MATH500 Learning Curves and Checkpoints}

The MATH results below retain all recorded checkpoints and the rank1 ablation. The extra update 110 continuation has a longer training budget and a changed learning-rate phase; it is supplementary to the update 90 main comparison.

\begin{figure}[H]
\centering
\includegraphics[width=\linewidth]{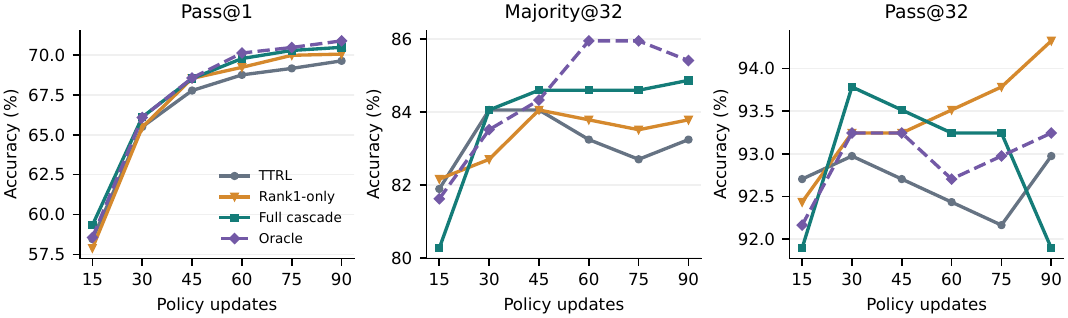}
\caption{\textbf{Reward intervention and policy learning.} TRUER moves Pass@1 and Majority@32 toward the ground-truth-reward reference, while Pass@32 shows a different ordering. Each line is one training seed. The figure shows all recorded checkpoints and does not select the best intermediate result.}
\label{fig:curves}
\end{figure}

\begin{figure}[H]
\centering
\includegraphics[width=\linewidth]{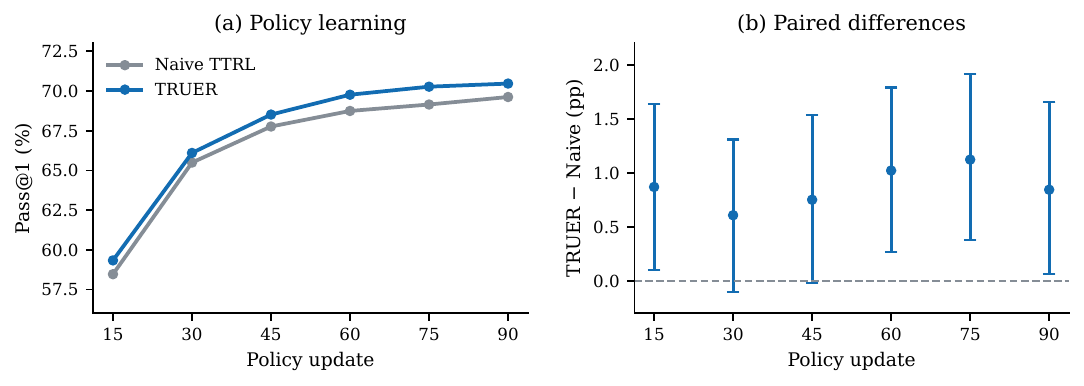}
\caption{\textbf{Paired uncertainty in the MATH500 learning curves.} Left: model Pass@1. Right: TRUER minus TTRL with pointwise 95\% question-bootstrap intervals, conditional on the saved samples and one training seed.}
\label{fig:traininguncertainty}
\end{figure}

Figure~\ref{fig:curves} compares the recorded learning curves, and Figure~\ref{fig:traininguncertainty} provides paired uncertainty estimates. Table~\ref{tab:allcheckpoints} lists the checkpoint scores; the rank1-only contrasts are reported in Appendix~\ref{app:rank1ablation}.

\begin{table}[H]
\centering
\caption{All recorded model checkpoints. Percentages, frozen scorer. No best-checkpoint selection is used for the main endpoint.}
\label{tab:allcheckpoints}
\small
\begin{tabular}{rlrrrrr}
\toprule
Step & Method & Greedy & Pass@1 & Pass@2 & Majority@32 & Pass@32 \\
\midrule
\inputtable{tables/all_steps_rows.tex}
\bottomrule
\end{tabular}
\end{table}

\subsection{AMC23 Learning Curves and Endpoints}
The following AMC figures use the seed 42 run with the 1.7B model and 4B rectifier reported in the main table. This run is distinct from the stable 1.7B-rectifier configuration. Only TRUER intermediate sampled checkpoints were retained; Naive and Oracle lines mark their endpoints.

\begin{figure}[H]
\centering
\includegraphics[width=\linewidth]{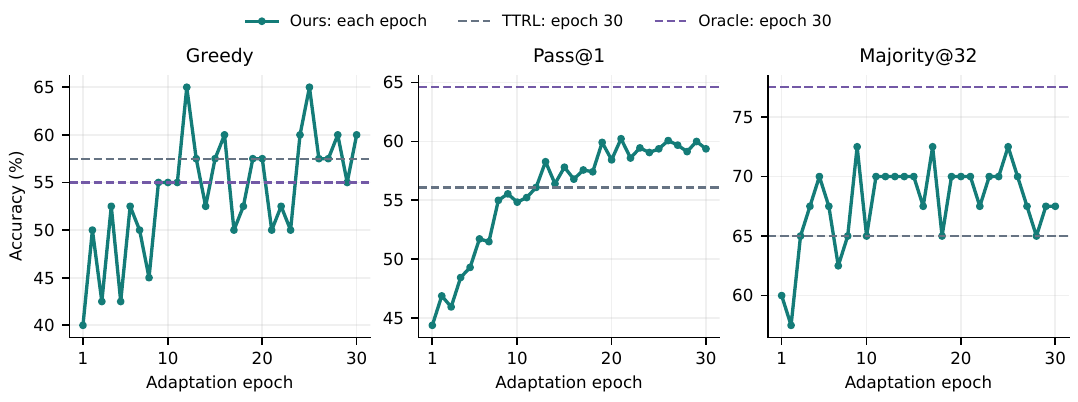}
\caption{\textbf{All 30 AMC23 TRUER checkpoints.} Solid curves are separately completed, matched evaluations of each checkpoint without rectifier-based selection. Dashed lines mark the final epoch 30 TTRL and Oracle scores; their intermediate sampled checkpoints were not retained. Greedy uses the same complete evaluation pipeline rather than the separate inline training validation.}
\label{fig:amccurves}
\end{figure}

Figure~\ref{fig:amccurves} shows the recorded learning curves; Table~\ref{tab:amcendpoints} reports the endpoint metrics.

\begin{table}[H]
\centering
\caption{AMC23 endpoints: 1.7B model, 4B rectifier, seed 42 (\%).}
\label{tab:amcendpoints}
\begin{tabular}{lrrrrr}
\toprule
Method & Greedy & Pass@1 & Pass@2 & Majority@32 & Pass@32 \\
\midrule
Base & 45.00 & 46.09 & 59.97 & 67.50 & 90.00 \\
TTRL & 57.50 & 56.09 & 61.67 & 65.00 & 77.50 \\
TRUER & 60.00 & 59.38 & 66.67 & 67.50 & 92.50 \\
Oracle & 55.00 & 64.61 & 72.91 & 77.50 & 92.50 \\
\bottomrule
\end{tabular}
\end{table}

\subsection{Answer-Distribution Changes}
\label{app:distribution}
\begin{figure}[H]
\centering
\includegraphics[width=\linewidth]{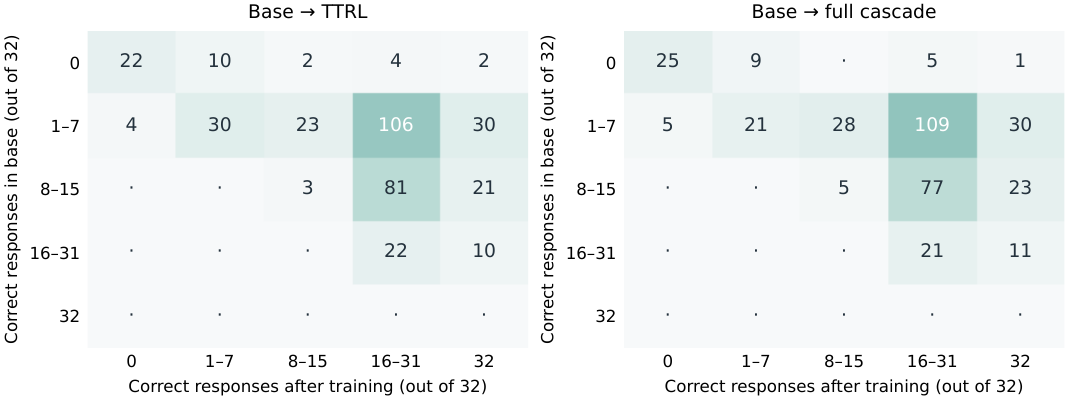}
\caption{\textbf{Where the same questions move after adaptation.} Rows are base correct-response counts; columns are counts at update 90. Both panels contain all 370 questions and use the same bins and color scale. Movement to the right indicates more frequent correct answers.}
\label{fig:transitions}
\end{figure}

The base model already samples a correct answer for 330 of the 370 MATH questions. Most additional correct samples after adaptation come from this already-covered set. Figure~\ref{fig:transitions} tracks each question's correct-answer count; no questions are selected by improvement. TRUER newly covers fifteen questions and loses five, alongside a larger increase in the frequency of correct answers. These are finite-sample coverage observations.

Table~\ref{tab:parsing} records parsing rates and generation lengths for the same endpoint comparison.

\begin{table}[H]
\centering
\caption{Parsing and generation lengths under the base comparison. All scores retain the full denominator of 11,840 responses; unparsed responses are not discarded.}
\label{tab:parsing}
\begin{tabular}{lrrr}
\toprule
Model & Unparsed & Length-limit hits & Mean output tokens \\
\midrule
Base & 32.97\% & 11.25\% & 1,102.7 \\
TTRL90 & 1.78\% & 2.58\% & 682.6 \\
TRUER90 & 1.98\% & 2.52\% & 684.1 \\
\bottomrule
\end{tabular}
\end{table}

\section{Reward Analysis and Ablations}
\label{app:reward}

\subsection{Fixed-Pool Reward Correction}
The fixed MATH TRUER trajectory contains 92,160 responses and 2,880 question visits. Majority labeling produces 4,012 FP and 1,088 FN rewards; rectification produces 1,081 FP and 2,284 FN rewards. Reward accuracy rises from 94.47\% to 96.35\%. TRUER retains 55,039 originally correct positive rewards, removes 1,557, and recovers 361, leaving 55,400 in total. Active groups fall from 2,557 to 2,105.

\begin{figure}[H]
\centering
\includegraphics[width=\linewidth]{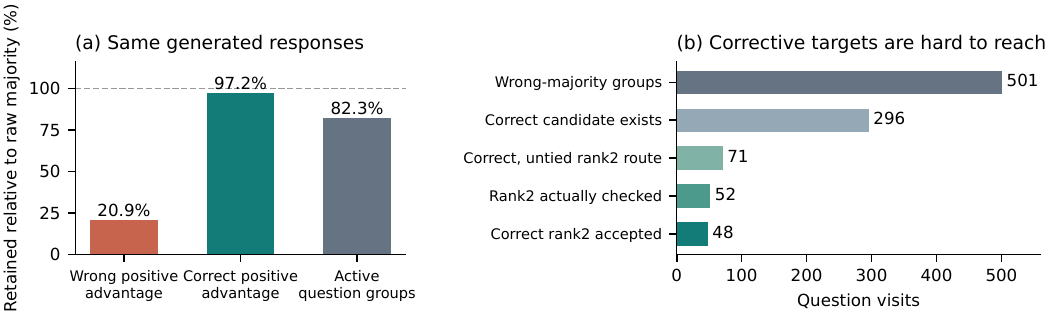}
\caption{\textbf{Cleaner rewards do not automatically supply more corrective signal.} Both panels use TRUER's fixed training response pool. Left: retained advantage mass and active groups relative to raw majority labeling. Right: the nested conditions needed to rescue a correct rank2. Counts are question visits, not unique questions.}
\label{fig:mechanism}
\end{figure}

At the question-visit level, 48 wrong labels become correct, 357 become abstentions, and 99 correct majorities are withheld. The rank2 branch accepts 48 correct and 25 wrong alternatives. Figure~\ref{fig:mechanism} relates these decisions to retained advantage mass and candidate availability. Counts are visits, not distinct questions.

Table~\ref{tab:trajectoryaudit} separately reports the full-run reward audit on each method's own responses.

\begin{table}[H]
\centering
\caption{Training reward audit over all 90 updates, evaluated on each arm's own responses. For an isolated labeling comparison, use the shared-pool numbers in the main text.}
\label{tab:trajectoryaudit}
\begin{tabular}{lrr}
\toprule
Quantity & TTRL & TRUER \\
\midrule
TP & 56,183 & 55,400 \\
FP & 4,363 & 1,081 \\
TN & 30,498 & 33,395 \\
FN & 1,116 & 2,284 \\
Overall reward accuracy & 94.05\% & 96.35\% \\
FPR & 12.52\% & 3.14\% \\
Positive-reward precision & 92.79\% & 98.09\% \\
Correct-response reward recall & 98.05\% & 96.04\% \\
Accepted question-group coverage & 100.00\% & 84.17\% \\
\bottomrule
\end{tabular}
\end{table}

\subsection{Comparison with Vote Filters}
\label{app:margin}
\begin{figure}[H]
\centering
\includegraphics[width=\linewidth]{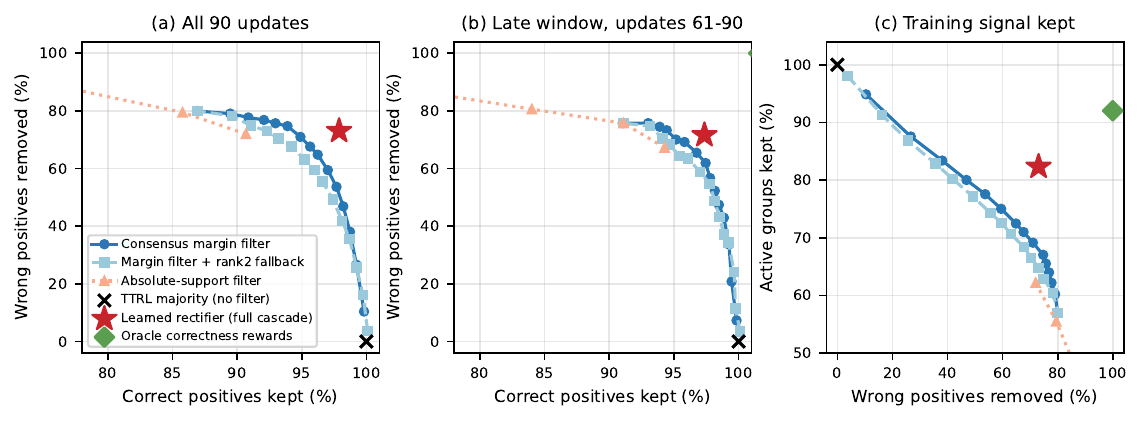}
\caption{\textbf{Label-free confidence filters against the learned rectifier.} Each curve sweeps one filter family over its threshold. At comparable FP removal, TRUER retains more correct rewards and active groups. Panel (b) restricts the pool to updates 61--90, where false positives dominate reward errors.}
\label{fig:margin}
\end{figure}

We replay vote-count and vote-margin thresholds on the same response pool (Table~\ref{tab:margin} and Figure~\ref{fig:margin}). At comparable FP removal, the rectifier retains more correct rewards and active groups than these frequency-only filters. The margin rule requires discarding about 2.7 times as many correct positive rewards to match TRUER's FP count. At aggressive thresholds it retains less wrong-positive advantage mass, but also loses more correct-positive mass.

\begin{table}[H]
\centering
\caption{Label-free confidence filters on the frozen pool of Section~\ref{sec:fixedpool}. Only vote counts enter the selection. Precision is $\mathrm{TP}/(\mathrm{TP}+\mathrm{FP})$; the last two columns are relative to unfiltered TTRL majority. The learned cascade removes 73.1\% of false positives while keeping 97.9\% of correct positive rewards and 82.3\% of active groups.}
{\small\setlength{\tabcolsep}{4pt}
\begin{tabular}{lrrrrrr}
\toprule
Policy & TP & FP & Active groups & Precision (\%) & TP kept (\%) & FP removed (\%) \\
\midrule
\inputtable{tables/margin_filter_rows.tex}
\bottomrule
\end{tabular}}
\label{tab:margin}
\end{table}

\subsection{Rank1-Only Ablation}
\label{app:rank1ablation}

Table~\ref{tab:rank1ablation} reports paired endpoint contrasts between TRUER, rank1-only, and naive TTRL. The checkpoint scores are listed in Table~\ref{tab:allcheckpoints}.

\begin{table}[H]
\centering
\caption{Paired 90-step contrasts involving rank1-only. Differences and intervals are percentage points. These were recomputed from the saved question-level scores after the rank1 run completed.}
\label{tab:rank1ablation}
\begin{tabular}{llrr}
\toprule
Contrast & Metric & Difference & 95\% interval \\
\midrule
\inputtable{tables/rank1_intervals_rows.tex}
\bottomrule
\end{tabular}
\end{table}

\subsection{Effect of Rectifier Training}

\label{app:verifiertraining}
\begin{figure}[H]
\centering
\includegraphics[width=\linewidth]{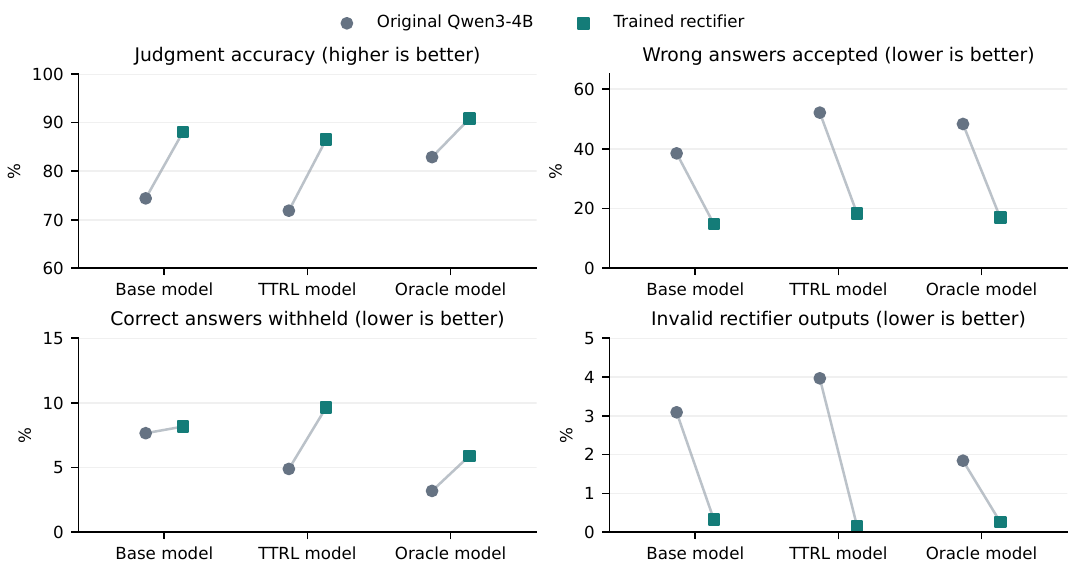}
\caption{\textbf{Task-specific training improves the 4B rectifier's judgment tradeoff.} Original and trained models judge the same candidates. Training reduces wrong-answer acceptance but also withholds more correct answers.}
\label{fig:verifiertraining}
\end{figure}

We compare the original Qwen3-4B and the trained rectifier on identical AMC23 candidate pools, using the same tokenized prompts, greedy decoding, parser, eligibility rules, and score threshold. The pools come from the base model, naive TTRL, and Oracle. Invalid outputs count as errors. Table~\ref{tab:verifiertraining} reports the numerical comparison, visualized in Figure~\ref{fig:verifiertraining}.

\begin{table}[H]
\centering
\caption{\textbf{Original versus trained rectifier on identical AMC23 responses.} Each entry is original$\to$trained, in percent. Accuracy includes invalid outputs as errors. Wrong-answer acceptance is conditional on a wrong candidate; correct-answer withholding includes explicit rejection and invalid outputs on correct candidates. The rectifier was trained on disjoint AMC2024 questions.}
\label{tab:verifiertraining}
{\small\setlength{\tabcolsep}{4pt}
\begin{tabular}{lrrrr}
\toprule
Response source & Accuracy $\uparrow$ & Wrong accepted $\downarrow$ & Correct withheld $\downarrow$ & Invalid $\downarrow$ \\
\midrule
Base model & $74.39\to88.05$ & $38.47\to14.80$ & $7.65\to8.16$ & $3.09\to0.33$ \\
TTRL model & $71.85\to86.52$ & $52.11\to18.35$ & $4.89\to9.64$ & $3.97\to0.16$ \\
Oracle model & $82.89\to90.79$ & $48.31\to16.92$ & $3.19\to5.89$ & $1.84\to0.26$ \\
\bottomrule
\end{tabular}}
\end{table}

Training improves judgment accuracy and reduces wrong-answer acceptance across these pools, with some increase in correct-answer withholding. Paired question-bootstrap intervals exclude zero for accuracy improvements and wrong-answer acceptance reductions. This control evaluates judgments; it does not isolate category weighting or replace a matched adaptation run using the untrained rectifier.

\paragraph{Adaptation with an untrained rectifier.}
Table~\ref{tab:untrained-adaptation} compares adaptation using the original pretrained rectifier and its trained counterpart. All three reported configurations start from the original Qwen3-1.7B policy and use seed 42, 120 policy updates, and the original endpoint evaluation with 32 responses per question. The trained-rectifier results use the original endpoints rather than the budget-control continuations. The trained rectifier yields higher final Pass@1 in each comparison; the untrained 1.7B rectifier on AIME24 also underperforms naive TTRL.

\begin{table}[H]
\centering\small
\caption{\textbf{Adaptation with and without rectifier training.} Final model Pass@1 (\%) without rectifier-based answer selection. All policies are Qwen3-1.7B; untrained denotes the original pretrained rectifier without task-specific training.}
\label{tab:untrained-adaptation}
\begin{tabular}{llrrr}
\toprule
Target & Rectifier & Naive TTRL & Untrained & Trained \\
\midrule
AMC23 & Qwen3-4B & 56.09 & 59.22 & \textbf{59.38} \\
AIME24 & Qwen3-1.7B & 20.21 & 18.96 & \textbf{21.46} \\
AIME24 & Qwen3-4B & 20.21 & 23.02 & \textbf{24.79} \\
\bottomrule
\end{tabular}
\end{table}

These comparisons have execution differences: the AMC23/4B and AIME24/4B untrained runs use policy data parallelism of four rather than two, while AIME24/1.7B changes from RTX 5090 with vLLM 0.23 to A800 with vLLM 0.25.1. They therefore do not isolate rectifier training under identical hardware and runtime settings.

\section{Case Studies}
\label{app:cases}
We select the successful example by the largest endpoint correct-sample gap among reviewed questions where naive TTRL is unanimously wrong and TRUER has at least 75\% correct responses. The failure example has the largest combined wrong-mode share among reviewed cases where both models have at least 75\% wrong consensus. Ties use question IDs. Known scorer errors and mixed-truth clusters are excluded.

\paragraph{Recovering a correct minority.}
For MATH algebra/2023, the question asks for the smallest integer in the domain of $\sqrt{x-1}/(x-2)$. The constraints are $x\geq1$ and $x\neq2$, so the answer is $1$; a common mistake gives $3$. TRUER rejects $3$ and selects the minority answer $1$ at updates 10, 13, and 34. Later visits have a correct majority. Endpoint samples contain 24 correct answers for TRUER, while naive TTRL answers $3$ in all 32 samples.

\begin{figure}[H]
\centering
\includegraphics[width=\linewidth]{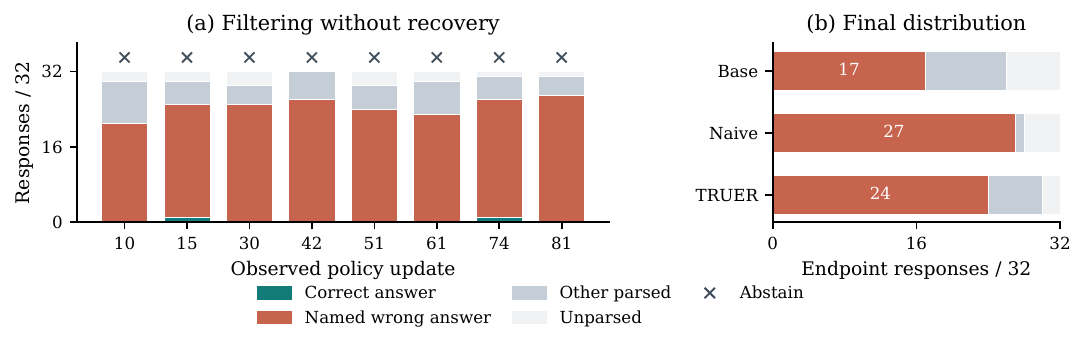}
\caption{\textbf{Filtering without recovery.} (a) Answer distributions at all eight recorded TRUER training visits; crosses mark abstention. (b) Separately sampled endpoint responses for Base, naive TTRL, and TRUER. The named wrong answer is zero. Rejecting this answer supplies no replacement target, and the model remains incorrect.}
\label{fig:filteringfailure}
\end{figure}

\paragraph{Filtering without recovery.}
For intermediate algebra/1279, summing the roots of $\sum_{j=1}^{4}j/(x^2-j)=2010x-4$ gives $2/1005$ by Vieta's formula after clearing denominators. Both models commonly answer zero. TRUER removes the wrong majority reward on all eight visits but abstains each time; the few correct candidates are blocked by rank2 tie rules. Neither model produces a correct endpoint sample (Figure~\ref{fig:filteringfailure}). Cleaner rewards can prevent an erroneous update without producing a corrective one.

\paragraph{An incorrect fallback.}
At update 64 on intermediate algebra/1849, the rectifier rejects rank1 answer $2$ but accepts rank2 answer $1$ with score 9. The true answer is $501$, absent from the candidate pool. The judgment misses a family of negative integer cases. Rejecting one wrong answer therefore does not certify its alternative.

\section{Deployment Selection and Cost}
\label{app:inference}
\label{app:amcinference}
\subsection{Shared Selection Rule}
The current deployment comparison uses one score-weighted answer aggregation rule for all six configurations. We retain each final model's original pool of 32 responses and apply the configuration's frozen rectifier to both naive TTRL and TRUER. A response is eligible when its final answer can be parsed and clustered, the model response is not truncated, and the rectifier prompt fits the 4,096-token input limit. Rectifier judgments use greedy decoding, a 4,096-token output cap, and an 8,192-token context. A valid integer score of at least 5 passes; malformed or truncated judgments do not pass.

For each answer $a$, let $\mathcal V_a$ contain the eligible candidates with that answer and a valid rectifier score $s_i$, and let $n_a=|\mathcal V_a|$. We compute
\begin{equation}
 S(a)=\frac{\sum_{i\in\mathcal V_a}\max(s_i-4,0)}{\sqrt{n_a}},\qquad n_a>0.
\label{eq:deploymentaggregation}
\end{equation}
Valid rejected judgments contribute zero to the numerator but remain in $n_a$; invalid judgments are excluded from both. The weights are used for ranking and are not calibrated probabilities. We select the answer with the largest positive support. Ties are resolved by the highest passing score in the answer cluster, then its original total candidate count, and finally the earliest index among its highest-scoring passing candidates. If no candidate passes, we return the original majority answer. Every question remains in the denominator.

\paragraph{Selector choice and scope.}
We selected this shared aggregation rule after comparing alternative selectors on the target response pools. The resulting deployment comparisons are exploratory and do not constitute independent validation of the selector choice. Candidate selection itself receives no target labels. The model checkpoints, rectifiers, candidate pools, eligibility criteria, and answer scorers remain fixed; this replay changes only how the saved judgments are aggregated. In particular, the AMC23 configuration with a 1.7B model and 4B rectifier uses the original seed 42 candidate pools and judgments.

Table~\ref{tab:unifieddeployment} reports the paired naive TTRL and TRUER results. The shared aggregation rule improves over each TRUER model's own majority voting in all six configurations. With matched rectifier postprocessing, TRUER is higher than naive TTRL in three configurations, tied in one, and lower in two. Historical selection controls below retain their original rules and results.

\begin{table}[H]
\centering\small
\caption{\textbf{Deployment under one shared selection rule (\%).} Parentheses identify the rectifier size. Both adaptation methods use the same frozen rectifier within each row. ``+Rectifier'' applies Equation~\ref{eq:deploymentaggregation} to the corresponding model's fixed 32-response pool, with majority fallback.}
\label{tab:unifieddeployment}
\setlength{\tabcolsep}{4pt}
\begin{tabular}{lr rrrr}
\toprule
& & \multicolumn{2}{c}{Naive TTRL} & \multicolumn{2}{c}{TRUER} \\
Model / target (rectifier) & $N$ & Majority & +Rectifier & Majority & +Rectifier \\
\midrule
Math-1.5B / MATH500 (4B) & 370 & 83.24 & 88.11 & 84.86 & 86.76 \\
Qwen3-1.7B / AMC23 (1.7B) & 40 & 65.00 & 65.00 & 67.50 & 70.00 \\
Qwen3-1.7B / AMC23 (4B) & 40 & 65.00 & 70.00 & 67.50 & 70.00 \\
Qwen3-1.7B / AIME24 (1.7B) & 30 & 26.67 & 30.00 & 26.67 & 33.33 \\
Qwen3-4B / AMC23 (4B) & 40 & 82.50 & 85.00 & 90.00 & 95.00 \\
Qwen3-4B / MATH-L5 (4B) & 600 & 84.83 & 87.17 & 86.00 & 86.17 \\
\bottomrule
\end{tabular}
\end{table}

\subsection{Historical Selection Controls}
Table~\ref{tab:amcinference} uses the same AMC1.7B/4B-rectifier seed 42 pools as the main table and the deployment comparison. It additionally tests top2 with one representative per cluster and top2 with all eligible responses in those clusters. These policies and the All32 selector rank passing candidates by score, cluster size, and sample order. None of the historical AMC selection gains establishes a statistically significant improvement under the recorded paired tests.

\begin{table}[H]
\centering
\caption{\textbf{Historical AMC23 final answer-selection accuracy (\%).} Every row uses the same 32 candidates for each model, and full question coverage. Changes do not alter the models' sampled Pass@1 or Pass@32.}
\label{tab:amcinference}
\begin{tabular}{lrrrr}
\toprule
Selection rule & Base & TTRL & TRUER & Oracle \\
\midrule
Original majority & 67.50 & 65.00 & 67.50 & 77.50 \\
Sequential rank1-to-rank2 & 72.50 & 67.50 & 67.50 & 80.00 \\
Top2: one representative & 72.50 & 67.50 & 65.00 & 80.00 \\
Top2: all candidates & 70.00 & 65.00 & 70.00 & 80.00 \\
All32: all candidates & 70.00 & 65.00 & 75.00 & 80.00 \\
\bottomrule
\end{tabular}
\end{table}

Table~\ref{tab:inference} reports the historical MATH fixed-pool selection controls, including the additional continuation described in Appendix~\ref{app:results}.

\begin{table}[H]
\centering
\caption{Historical MATH fixed-pool selection controls (percent). Oracle receives matched postprocessing here.}
\label{tab:inference}
\begin{tabular}{lrrr}
\toprule
Selection rule & TRUER90 & TRUER110 & Oracle90 \\
\midrule
Original majority & 84.86 & 85.14 & 85.41 \\
Rank1-to-rank2 cascade & 85.68 & 86.22 & 86.22 \\
Top2, score first & 85.68 & 85.95 & 86.76 \\
Top3, score first & 85.68 & 85.68 & 87.84 \\
\bottomrule
\end{tabular}
\end{table}

\subsection{Inference Cost}
The weighted aggregation rule reuses the same at most 32 candidate judgments per question as All32. No new judgments are generated for this replay. Table~\ref{tab:unifiedcost} records the judgments and rectifier output tokens for the current paired evaluations; these totals exclude prompt tokens and the unchanged response generation.

\begin{table}[H]
\centering\small
\caption{\textbf{Rectifier inference cost for the unified deployment comparison.} Counts cover all eligible responses in each fixed pool.}
\label{tab:unifiedcost}
\setlength{\tabcolsep}{4pt}
\begin{tabular}{lrrrr}
\toprule
& \multicolumn{2}{c}{Naive TTRL} & \multicolumn{2}{c}{TRUER} \\
Model / target (rectifier) & Judgments & Output tokens & Judgments & Output tokens \\
\midrule
Math-1.5B / MATH500 (4B) & 11,463 & 7,409,697 & 11,457 & 7,382,694 \\
Qwen3-1.7B / AMC23 (1.7B) & 1,261 & 1,501,618 & 1,207 & 1,366,736 \\
Qwen3-1.7B / AMC23 (4B) & 1,261 & 1,110,284 & 1,173 & 939,509 \\
Qwen3-1.7B / AIME24 (1.7B) & 864 & 208,525 & 768 & 169,730 \\
Qwen3-4B / AMC23 (4B) & 1,194 & 936,227 & 1,170 & 925,414 \\
Qwen3-4B / MATH-L5 (4B) & 17,801 & 11,997,996 & 17,359 & 8,656,022 \\
\bottomrule
\end{tabular}
\end{table}

\begin{figure}[H]
\centering
\includegraphics[width=\linewidth]{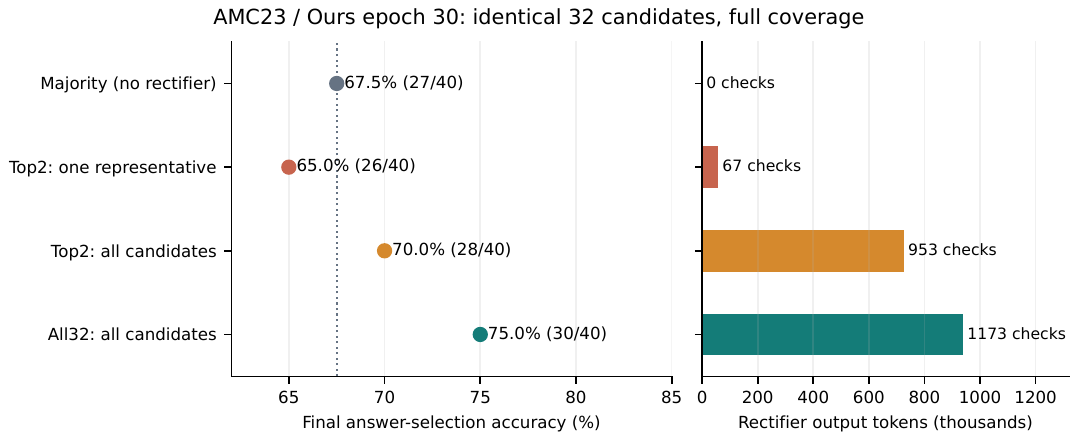}
\caption{\textbf{Selection quality and additional rectifier cost on AMC23.} The same frozen rectifier scores TRUER epoch 30 candidates. Output-token counts are sums from the cached judgments, excluding rectifier input tokens and the unchanged response generation. Calls count candidate judgments. All policies retain all 40 questions.}
\label{fig:amcinference}
\end{figure}

On historical AMC TRUER candidates, top2 representatives require 67 judgments and 58,074 rectifier output tokens, while All32 requires 1,173 judgments and 939,509 output tokens. These counts exclude input tokens and response generation. Figure~\ref{fig:amcinference} shows the recorded tradeoff.

\section{Training-Budget Controls}
\label{app:budgetcontrols}
We conduct two independent continuation controls on configurations from the main table. The GPU-hour control extends naive TTRL and compares it with the original TRUER checkpoint. The active-group control extends TRUER and compares it with the original naive TTRL checkpoint; we also report its change from the original TRUER endpoint. The two extended endpoints do not simultaneously match computation and effective training exposure.

\paragraph{Training overhead.}
The recorded MATH policy updates take approximately 219.3 seconds for TRUER versus 177.2 for naive TTRL, with a dedicated rectifier GPU; this excludes rectifier training. Matching policy updates is not total-compute matching.

\subsection{GPU-Hour-Matched Naive TTRL}
Each run starts from the original naive TTRL endpoint and continues until its cumulative adaptation-stage GPU-hours approximately match those of the original TRUER run. The stopping budget is fixed before endpoint evaluation. Table~\ref{tab:budgetcompute} reports all five configurations. The final naive budgets are 1.1--1.4\% above the corresponding TRUER budgets, reflecting the budget margin and stopping at complete updates. Additional naive training improves Pass@1 in three configurations and reduces it in two; the original TRUER checkpoints remain higher in all five comparisons.

\begin{table}[H]
\centering\small
\caption{\textbf{GPU-hour-matched naive continuation.} $N_0$ and $T_0$ are the original naive TTRL and TRUER endpoints; $N_+$ is the extended naive TTRL endpoint. GPU-hours cover adaptation and online rectifier inference, excluding rectifier pretraining, saving, export, and evaluation.}
\label{tab:budgetcompute}
\setlength{\tabcolsep}{3pt}
\begin{tabular}{lrrrrrrr}
\toprule
& & \multicolumn{3}{c}{Pass@1 (\%)} & \multicolumn{3}{c}{GPU-hours} \\
\cmidrule(lr){3-5}\cmidrule(lr){6-8}
Model / task & Seed & $N_0$ & $N_+$ & $T_0$ & $N_0$ & $N_+$ & $T_0$ \\
\midrule
Qwen3-1.7B / AMC23 & 42 & 56.09 & 59.14 & 59.38 & 9.78 & 27.65 & 27.31 \\
Qwen3-1.7B / AIME24 & 42 & 20.21 & 23.75 & 24.79 & 19.70 & 39.39 & 38.96 \\
Qwen3-4B / AMC23 & 43 & 79.69 & 78.52 & 82.34 & 27.10 & 30.64 & 30.21 \\
Qwen3-4B / MATH-L5 & 43 & 79.82 & 80.36 & 80.68 & 56.45 & 68.75 & 67.84 \\
Llama-3.1-8B / AMC23 & 43 & 49.45 & 47.50 & 60.23 & 29.26 & 63.69 & 62.97 \\
\bottomrule
\end{tabular}
\end{table}

\subsection{Active-Group-Matched TRUER}
An active group contains 32 responses with nonconstant binary rewards and hence nonzero group-relative outcome advantages. We extend each original TRUER checkpoint until its cumulative active-group count reaches that of the original naive TTRL run, stopping at the first complete update that crosses the target. This matches effective group exposure, not optimizer-step counts or accumulated parameter-gradient norms. Counts are independently checked against the saved per-response rewards at every update. Table~\ref{tab:budgetactive} reports the active-group-matched results.

\begin{table}[H]
\centering\small
\caption{\textbf{Active-group-matched TRUER continuation.} $T_+$ is the extended TRUER endpoint. Group counts include the original training; the target is the original naive TTRL count. GPU-hours are cumulative. Seeds and original scores are given in Table~\ref{tab:budgetcompute}.}
\label{tab:budgetactive}
\setlength{\tabcolsep}{3pt}
\begin{tabular}{lrrrrrr}
\toprule
& \multicolumn{3}{c}{Active groups} & & & \\
\cmidrule(lr){2-4}
Model / task & $T_0$ & Target & $T_+$ & Added steps & Pass@1 (\%) & GPU-hours \\
\midrule
Qwen3-1.7B / AMC23 & 636 & 930 & 935 & 65 & 61.02 & 43.46 \\
Qwen3-1.7B / AIME24 & 530 & 1,179 & 1,180 & 144 & 27.92 & 121.80 \\
Qwen3-4B / AMC23 & 569 & 651 & 653 & 21 & 81.72 & 38.00 \\
Qwen3-4B / MATH-L5 & 1,094 & 1,197 & 1,197 & 24 & 80.74 & 76.15 \\
Llama-3.1-8B / AMC23 & 415 & 630 & 630 & 139 & 64.53 & 140.13$^{\dagger}$ \\
\bottomrule
\end{tabular}
\end{table}

$^{\dagger}$The Llama GPU-hour entry is a lower bound; the recorded cumulative cost is 140.13--140.60 GPU-hours, including discarded computation after checkpoint recovery. Active-group counts include only the retained training trajectory.

All five active-group-matched endpoints remain above the original naive TTRL scores. Relative to the original TRUER checkpoints, Pass@1 improves in four configurations, with gains of 3.13 percentage points on Qwen3-1.7B/AIME24 and 4.30 points on Llama/AMC23. Qwen3-4B/AMC23 decreases by 0.63 points, while Qwen3-4B/MATH-L5 changes little ($+0.06$ points). Thus, additional effective training benefits most configurations, but does not uniformly improve TRUER.

\subsection{Continuation and Evaluation Protocols}
Both controls restart Adam from the corresponding final model weights while retaining the original base reference model. The TRUER continuations retain the frozen rectifier. Budgets determine stopping independently of evaluation scores. Endpoint evaluation retains each original configuration's questions, scoring rules, and full 32-sample plus greedy protocol. GPU-hours count active training time multiplied by the participating physical GPUs, including online rectifier services and counting shared GPUs only once. They exclude rectifier pretraining, checkpoint saving, export, and endpoint evaluation. Hardware allocations and parallel topologies differ in some continuations, so these are approximate adaptation-compute controls rather than hardware-normalized FLOP comparisons or from-scratch runs under a common total budget.

These controls use one training seed per configuration. Conditional question-bootstrap 95\% intervals for the TRUER-minus-extended-naive Pass@1 difference are $[0.94,7.03]$ points on Qwen3-4B/AMC23, $[-10.23,11.17]$ on Qwen3-1.7B/AMC23, and $[-1.88,27.81]$ on Llama/AMC23. For extended-minus-original TRUER, the corresponding available intervals are $[-0.55,3.67]$ on Qwen3-1.7B/AMC23 and $[-2.19,1.09]$ on Qwen3-4B/AMC23. These intervals condition on the retained runs and do not include training-seed variation.

\section{Reproducibility}
Figures and tables use retained question-level scores, reward audits, and response histories. Main-table data, reward-window counts, figure inputs, and source hashes are preserved with the analysis. Historical configurations and exploratory selection comparisons are identified separately. The continuation controls additionally retain budget ledgers and complete endpoint evaluations.

\end{document}